\documentclass{article}
\usepackage{spconf}
\usepackage{cite}
\usepackage{amsmath,amssymb,amsfonts}
\usepackage{algorithmic}
\usepackage{graphicx}
\usepackage{textcomp}
\usepackage{xcolor}
\usepackage{balance}
\usepackage{acronym}
\usepackage{hyperref}
\usepackage{csquotes}
\usepackage{adjustbox}
\usepackage{nicematrix}
\usepackage{booktabs}
\usepackage{multirow}
\usepackage{siunitx}
\usepackage{tikz,pgfplots}
\usepackage[capitalize,nameinlink]{cleveref}
\usepackage{enumitem}

\crefname{figure}{Fig.\@}{Figs.\@}
\Crefname{figure}{Figure\@}{Figures\@}

\crefname{table}{Table\@}{Tables\@}
\Crefname{table}{Table\@}{Tables\@}
\pgfplotsset{compat=newest}
\usepgfplotslibrary{colorbrewer}
\usepgfplotslibrary{fillbetween}
\usepgfplotslibrary{groupplots}

\makeatletter
\newcommand*{\org@overidelabel}{}
\let\org@overridelabel\AC@verridelabel
\renewcommand*{\AC@verridelabel}[1]{%
  \@bsphack
  \protected@write\@auxout{}{\string\AC@undonewlabel{#1@cref}}%
  \org@overridelabel{#1}%
  \@esphack
}%
\makeatother

\newcommand{\BE}{\begin{equation}\begin{aligned}}
\newcommand{\EE}{\end{aligned}\end{equation}}

\DeclareMathOperator*{\argmin}{arg\,min}

\DeclareSIUnit{\decibel}{dB}

\acrodef{asd}[ASD]{anomalous sound detection}
\acrodef{knn}[KNN]{k-nearest neighbors}
\acrodef{nn}[NN]{nearest neighbor}
\acrodef{ldn}[LDN]{local density-based normalization}
\acrodef{varmin}[VarMin]{variance minimization}

\begin{document}
\ninept
\title{Anomaly-Free Self-Optimization via AUC Bounds}

\name{Kevin Wilkinghoff$~^{1,2}$, Zheng-Hua Tan$^{1,2}$}
\address{$^{1}$Department of Electronic Systems, Aalborg University, Denmark, $^{2}$Pioneer Centre for AI, Denmark}

\maketitle

\begin{abstract}
Anomalies are rare, and anomalous data are often unavailable during development, making it difficult to determine which anomaly detection models and configurations will generalize to unseen anomalies. Recent approaches address this challenge by generating pseudo-anomalies and using bounds on the achievable area under the ROC curve (AUC) to select the optimal configuration from a finite set of candidates. Instead, we use the AUC bound as a differentiable, anomaly-free objective for directly optimizing continuous parameters of anomaly detection systems. We demonstrate this framework by optimizing ensemble weights and introducing a learnable score-rescaling mechanism that adapts pseudo-anomaly scores, enabling optimization beyond a predefined candidate set. Experiments across multiple datasets and embedding models show that AUC-bound optimization achieves significant performance gains over conventional model selection and prior development-set-based parameter selection. The results further show that direct optimization is less sensitive to the choice of pseudo-anomaly construction.
\end{abstract}

\begin{keywords}
anomaly detection,
anomaly-free optimization,
self-optimization,
ensemble learning,
pseudo anomalies
\end{keywords}

\acresetall

\section{Introduction}

\Ac{asd} systems are typically trained without anomalous data because
anomalies are rare and difficult to obtain. However, developing such systems still requires selecting appropriate models and system parameters, including hyperparameters, pooling strategies, and ensemble weights. Conventional approaches often perform this tuning using anomalous development data \cite{dohi2023description,nishida2024description}, creating a dependence on anomalous data even when the underlying model is trained entirely on normal data. Such data may be unavailable, scarce, or unreliable, particularly when the anomalies of interest are unknown during system development. Importantly, this dependence extends beyond methods that explicitly train on anomalous data. Even training-free \ac{asd} approaches often tune hyperparameters according to performance on an anomalous development set \cite{guan2023time-weighted,saengthong2025deep,wilkinghoff2026temporal}. These challenges motivate approaches that can adapt \ac{asd} systems without relying on anomalous data.
\par
Anomaly-free model selection addresses this dependence by replacing real anomalies with pseudo-anomalies. In particular, prior work has shown that an AUC-derived bound can be used to estimate relative anomaly detection performance from normal data and pseudo-anomalies, enabling model selection without anomalous development labels \cite{wilkinghoff2026can}. However, the resulting selection performance can depend strongly on the quality of the pseudo-anomalies, and the strongest-performing constructions rely on labeled cross-domain, cross-class, or cross-attribute data. Moreover, this use of the bound remains fundamentally selection-based: a finite set of candidate configurations must first be evaluated, after which the candidate with the most favorable bound is selected. As the number of tunable parameters increases, such a search can become increasingly costly and requires continuous parameters to be discretized into a predefined set of candidates.
\par
In this work, we use the AUC-derived bound as a differentiable optimization
objective rather than only as a criterion for ranking predefined
configurations. This enables anomaly-free self-optimization, that is, direct optimization of continuous \ac{asd}-system parameters without anomalous development labels or discretizing continuous parameters into a predefined candidate set. We demonstrate this principle by optimizing ensemble weights and introduce a learnable pseudo-anomaly re-scaling mechanism to adapt the surrogate outlier distribution to the scale relevant for optimization.

The main contributions of this work are:
\begin{itemize}[noitemsep, topsep=0pt]
    \item \textbf{Anomaly-free self-optimization:} We show that an AUC-derived bound can be optimized directly to tune \ac{asd}-system parameters without anomalous development labels, with ensemble weights as a concrete example.
    \item \textbf{Pseudo-anomaly re-scaling:} We introduce a learnable
    re-scaling mechanism that adapts the surrogate pseudo-anomaly distribution to the scale relevant for optimization.
    \item \textbf{Empirical gains:} Across the evaluated \ac{asd} settings, the proposed approach improves over conventional model selection and prior development-set parameter selection, with the best performance obtained using synthetic pseudo-anomalies that require no metadata and can be generated directly from normal reference data.
\end{itemize}

\section{Related Work}

\textbf{Ensemble learning for \ac{asd}:}
Ensemble methods improve \ac{asd} by combining information from multiple models, either at the representation or score level. At the representation level, multiple embedding models can be combined before anomaly scoring \cite{fujimura2025discriminative}. At the score level, previous approaches have summed standardized anomaly scores from different models \cite{giri2020self,kuroyanagi2021ensemble}, different input features
\cite{kawaguchi2019anomaly}, or models with different parameter settings \cite{wilkinghoff2021sub-cluster}. Other methods learn convex combinations of anomaly scores to account for the strengths and complementary information provided by individual models, with the ensemble weights optimized using anomalous development data \cite{lopez2021ensemble,deng2022ensemble}. In contrast, our approach learns continuous ensemble weights without access to anomalies.

\textbf{\ac{asd} model selection:}
Unsupervised model selection has been explored through internal evaluation criteria based on input data and model scores \cite{ma2023need}, meta-learning from historical labeled tasks \cite{zhao2021automatic}, and synthetic anomalies for constructing surrogate validation sets \cite{fung2025model}. More recently, AutoUAD uses anomaly-free surrogate metrics together with Bayesian optimization for hyperparameter tuning \cite{dai2025autouad}. Our prior work introduced an AUC-derived bound with pseudo-anomalies for anomaly-free model selection \cite{wilkinghoff2026can}. In contrast, the present work uses this bound as a differentiable objective for directly optimizing \ac{asd} system parameters.

\section{Anomaly-Free AUC-Bound Optimization}

\subsection{AUC-bound optimization}

Let $z(x;\theta)$ denote an anomaly score for a sample $x$ parameterized by system parameters $\theta$. Given a reference set of normal samples, we obtain the inlier score distribution $z_{\mathrm{in}}(\theta)$ from within-reference-set distances, while $z_{\mathrm{out}}(\theta)$ denotes scores from pseudo-anomalies generated from the same normal data. We use the AUC-derived bound introduced in \cite{wilkinghoff2026can} as the anomaly-free optimization objective
\begin{equation}
\mathcal{B}(\theta)
=
1+
\frac{
\operatorname{Var}[z_{\mathrm{in}}(\theta)]
+
\operatorname{Var}[z_{\mathrm{out}}(\theta)]
}{
\left(
\mathbb{E}[z_{\mathrm{out}}(\theta)]
-
\mathbb{E}[z_{\mathrm{in}}(\theta)]
\right)^2
}
\end{equation}
where $\operatorname{AUC}\geq1/\mathcal{B}(\theta)$, so minimizing
$\mathcal{B}(\theta)$ maximizes the corresponding lower bound on the
AUC. Although the bound is defined in terms of the score distributions of normal and anomalous samples, the anomalous distribution is unavailable during anomaly-free optimization. Following \cite{wilkinghoff2026can}, we replace real anomalies with pseudo-anomalies generated from normal data. Given $M$ candidate \ac{asd} systems with parameters $\Theta=\{\theta_1,\ldots,\theta_M\}$ and a fixed normal reference set and pseudo-anomaly set, previous work selects the candidate with the smallest bound,
\begin{equation}
m^*
=
\argmin_{m=1,\ldots,M}
\mathcal{B}(\theta_m).
\end{equation}
Instead of selecting from a finite set of candidates, we optimize the bound directly over the continuous parameter space,
\begin{equation}
\theta^*
=
\argmin_{\theta}
\mathcal{B}(\theta).
\end{equation}
This is possible because the AUC bound, $\mathcal{B}$, depends only on the first and second moments of the score distributions, which can be computed and differentiated with respect to the system parameters. In contrast, direct optimization of the pseudo-AUC, computed from normal samples and pseudo-anomalies, depends on pairwise score comparisons and ranking operations, making direct gradient-based optimization difficult. Thus, when pseudo-anomalies can be generated from normal data without auxiliary metadata, $\mathcal{B}$ provides a differentiable objective for self-optimization, enabling ASD-system parameters to be optimized without prior knowledge of the anomaly distribution or access to anomalous development samples.

\subsection{Anomaly-free ensemble learning}

As a concrete instance of the proposed self-optimization framework, we consider an ensemble of $M$ candidate anomaly scores $z_m(x)$,
\begin{equation}
z(x)
=
\sum_{m=1}^{M} w_m z_m(x),
\end{equation}
where
\begin{equation}
w_m
=
\frac{\exp(a_m)}
{\sum_{j=1}^{M}\exp(a_j)}.
\end{equation}
With the candidate system parameters fixed, the ensemble
weights $\mathbf{a}$ are optimized by minimizing the AUC-bound objective
\begin{equation}
\mathbf{a}^*
=
\argmin_{\mathbf{a}}
\mathcal{B}\!\left(
z_{\mathrm{in}}(\mathbf{a}),
z_{\mathrm{out}}(\mathbf{a})
\right),
\end{equation}
where
\begin{equation}
z_{\mathrm{in}}(\mathbf{a})
=
\sum_{m=1}^{M} w_m(\mathbf{a})z_{m,\mathrm{in}},
\qquad
z_{\mathrm{out}}(\mathbf{a})
=
\sum_{m=1}^{M} w_m(\mathbf{a})z_{m,\mathrm{out}}.
\end{equation}
The initialization $a_m=0$ yields equal weights, whereas a one-hot weight vector corresponds to selecting a single candidate. The formulation therefore generalizes both equal-weight ensembling and single-model selection.

\subsection{Adaptive pseudo-outlier score scaling}
\begin{figure}[t]
\centering
\begin{adjustbox}{max width=\columnwidth}
\includegraphics[width=\linewidth]{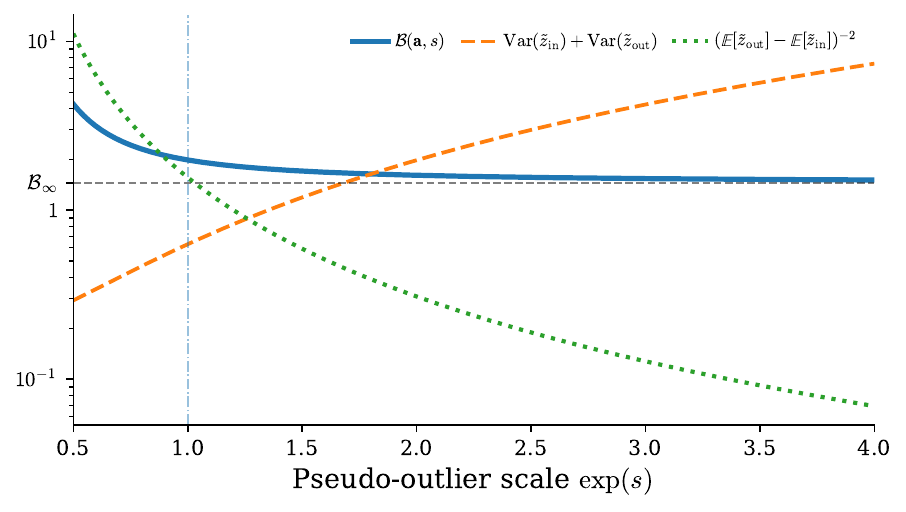}
\end{adjustbox}

\caption{Effect of the pseudo-outlier scale $\alpha=\exp(s)$ on the AUC-derived bound. Increasing $\alpha$ increases the separation between inlier and pseudo-outlier scores, but also the pseudo-outlier variance. The bound decreases before approaching a limiting value.}
\label{fig:scale_balancing}

\end{figure}

Pseudo-outlier constructions can produce different score scales across
candidate models, which can lead to poorly calibrated pseudo-outlier and
inlier scores and distort comparisons between candidates. We therefore
introduce a positive learnable scale $\alpha=\exp(s)$ for the pseudo-outlier scores. For log-distance scores, we apply $\operatorname{softplus}$ to obtain
positive scores and scale only the pseudo-outliers,
\begin{equation}
\tilde z_{\mathrm{in}}
=
\operatorname{softplus}(z_{\mathrm{in}}),
\qquad
\tilde z_{\mathrm{out}}
=
\alpha\cdot\operatorname{softplus}(z_{\mathrm{out}}).
\end{equation}
We use the bound on these transformed scores as a surrogate objective and jointly optimize \(s\) and the ensemble parameters by
\begin{equation}
\min_{\mathbf{a},s}
\mathcal{B}\!\left(
\tilde z_{\mathrm{in}}(\mathbf{a}),
\tilde z_{\mathrm{out}}(\mathbf{a})
\right).
\end{equation}
The scale is used only for evaluating the anomaly-free objective and does not modify the final anomaly scores used for evaluation. Increasing $\alpha$ increases both the pseudo-outlier variance and the mean separation in the AUC bound, while the resulting improvement diminishes as $\alpha$ grows, as illustrated in \cref{fig:scale_balancing}.

\section{Experimental Setup}

\begin{table*}[t]
\centering
\vspace{-6.2pt}
\caption{Comparison of \ac{asd} system selection and ensemble strategies. Entries report the DCASE benchmark metrics (\%) and their average.}
\label{tab:ensemble_main}

\setlength{\tabcolsep}{3pt}
\begin{adjustbox}{max width=0.95\textwidth}
\begin{NiceTabular}{ll*{11}{c}}
\toprule

&
&
\multicolumn{2}{c}{requires labels for}
&
\multicolumn{2}{c}{DCASE~2022}
&
\multicolumn{2}{c}{DCASE~2023}
&
\multicolumn{2}{c}{DCASE~2024}
&
\multicolumn{2}{c}{DCASE~2025}
&

\\

\cmidrule(lr){3-4}
\cmidrule(lr){5-6}
\cmidrule(lr){7-8}
\cmidrule(lr){9-10}
\cmidrule(lr){11-12}

Scoring Paradigm
&
Method
&
Anomalies
&
Metadata
&
\multicolumn{1}{c}{Dev.}
&
\multicolumn{1}{c}{Eval.}
&
\multicolumn{1}{c}{Dev.}
&
\multicolumn{1}{c}{Eval.}
&
\multicolumn{1}{c}{Dev.}
&
\multicolumn{1}{c}{Eval.}
&
\multicolumn{1}{c}{Dev.}
&
\multicolumn{1}{c}{Eval.}
&
\multicolumn{1}{c}{Avg.}
\\

\midrule

\multirow{8}{*}{\acs{nn}}
& Oracle-Selected&\checkmark&--
& 69.11 & 66.63 & 65.65 & 67.41 & 63.23 & 63.28 & 66.63 & 62.17 & 65.51
\\
& Fixed-Selected&\checkmark&--
& \pmb{65.71} & \underline{62.95} & \pmb{64.14} & 62.24 & \pmb{59.62} & 57.92 & \pmb{63.06} & \underline{58.30} & \pmb{61.74}
\\
& Random-Selected&--&--
& 61.46 & 59.20 & 60.20 & 60.69 & 57.01 & 56.63 & 60.18 & 55.54 & 58.86
\\
& Equal&--&--
& 62.11 & 59.83 & 61.07 & 61.30 & 58.10 & 57.36 & 61.76 & 56.33 & 59.73
\\
& Pseudo-AUC-Selected (Diverse) \cite{wilkinghoff2026can}&--&\checkmark
& \underline{65.69} & \pmb{63.19} & 60.02 & \pmb{66.01} & \underline{59.19} & \pmb{58.79} & 61.58 & 56.65 & \underline{61.39}
\\
& Bound-Selected (Diverse) \cite{wilkinghoff2026can}&--&\checkmark
& 64.07 & 62.21 & \underline{61.79} & 63.02 & 57.20 & 58.30 & 62.30 & 57.64 & 60.82
\\
& Proposed Approach (Feature)&--&--
& 62.80 & 60.85 & 61.76 & \underline{64.30} & 58.84 & \underline{58.61} & \underline{62.59} & \pmb{58.61} & 61.05
\\

\midrule

\multirow{8}{*}{\ac{nn} + \acs{ldn} \cite{wilkinghoff2025local}}
& Oracle-Selected&\checkmark&--
& 70.98 & 70.30 & 70.25 & 72.74 & 63.50 & 67.45 & 65.09 & 66.49 & 68.35
\\
& Fixed-Selected&\checkmark&--
& \pmb{66.21} & 63.98 & \pmb{66.15} & \underline{69.34} & 58.94 & 60.79 & \underline{61.35} & 61.43 & 63.52
\\
& Random-Selected&--&--
& 62.39 & 62.44 & 62.22 & 64.61 & 57.27 & 59.21 & 59.52 & 59.56 & 60.90
\\
& Equal&--&--
& 64.56 & 65.03 & 64.80 & 69.02 & 59.22 & \underline{61.51} & \pmb{61.69} & \underline{62.87} & \underline{63.59}
\\
& Pseudo-AUC-Selected (Diverse) \cite{wilkinghoff2026can}&--&\checkmark
& 65.30 & \underline{65.14} & 62.72 & 69.13 & \underline{59.38} & 57.96 & 58.22 & 62.01 & 62.48
\\
& Bound-Selected (Diverse) \cite{wilkinghoff2026can}&--&\checkmark
& 63.98 & 63.10 & 60.68 & 67.12 & 58.04 & 60.96 & 61.15 & 62.32 & 62.17
\\
& Proposed Approach (Feature)&--&--
& \underline{65.90} & \pmb{65.88} & \underline{66.04} & \pmb{70.43} & \pmb{59.66} & \pmb{62.26} & 61.20 & \pmb{63.95} & \pmb{64.41}
\\

\midrule

\multirow{8}{*}{\acs{nn} + \acs{ldn} + VarMin \cite{matsumoto2025adjusting}}
& Oracle-Selected&\checkmark&--
& 71.96 & 70.69 & 70.13 & 73.03 & 64.32 & 67.17 & 66.56 & 66.19 & 68.76
\\
& Fixed-Selected&\checkmark&--
& \underline{66.91} & 65.20 & \pmb{66.42} & 69.46 & 59.63 & 60.47 & \underline{63.59} & 62.06 & \underline{64.22}
\\
& Random-Selected&--&--
& 63.22 & 63.19 & 62.94 & 65.70 & 57.86 & 59.70 & 61.32 & 59.02 & 61.62
\\
& Equal&--&--
& 65.31 & 65.34 & 65.43 & 69.38 & \underline{59.75} & \underline{62.10} & 63.67 & 61.75 & 64.09
\\
& Pseudo-AUC-Selected (Diverse)\cite{wilkinghoff2026can}&--&\checkmark
& \pmb{67.33} & \pmb{66.82} & 61.91 & \pmb{72.04} & 59.63 & 60.42 & 60.91 & 61.66 & 63.84
\\
& Bound-Selected (Diverse) \cite{wilkinghoff2026can}&--&\checkmark
& 64.42 & 65.11 & 62.19 & 70.17 & 57.60 & 61.02 & 62.43 & \underline{62.85} & 63.22
\\
& Proposed Approach (Feature)&--&--
& 66.15 & \underline{66.14} & \underline{66.20} & \underline{71.12} & \pmb{60.10} & \pmb{62.44} & \pmb{63.81} & \pmb{63.16} & \pmb{64.89}
\\

\bottomrule
\end{NiceTabular}
\end{adjustbox}
\end{table*}

\subsection{Datasets and candidate models}
We evaluate the proposed approach on the DCASE~2022--2025 benchmark datasets \cite{dohi2022description,dohi2023description,nishida2024description,nishida2025description}, covering MIMII-DG \cite{dohi2022mimiidg}, ToyADMOS2 \cite{harada2021toyadmos2}, ToyADMOS2+ \cite{harada2023toyadmos2+}, ToyADMOS2\# \cite{niizumi2024toyadmos2sharp}, ToyADMOS2025 \cite{harada2025toyadmos2025}, and IMAD-DS \cite{albertini2024imadds}. We use the official DCASE metric, defined as the harmonic mean of source-domain AUC, target-domain AUC, and joint pAUC ($p=0.1$). We report 95\% confidence intervals for pairwise differences based on paired bootstrap resampling. Ensemble parameters are optimized with Adam for 100 steps at a learning rate of 0.05, starting from zero logits and log-scale, yielding uniform weights and an initial pseudo-outlier scale of one.
\par
We consider four pre-trained self-supervised audio embedding models: OpenL3 \cite{cramer2019look}, BEATs \cite{chen2023beats}, EAT \cite{chen2024eat}, and Dasheng \cite{dinkel2024dasheng}, each combined with 52 pooling configurations comprising mean, max, generalized mean pooling \cite{radenovic2019fine-tuning} with $p=1,\ldots,25$, and relative deviation pooling \cite{wilkinghoff2026temporal} with $\gamma=1,\ldots,25$. This results in 208 candidate systems. Anomaly scores are computed using the logarithmic \ac{nn} distance, with or without \ac{ldn} \cite{wilkinghoff2025local} ($K=2$), and with or without its \ac{varmin} extension \cite{matsumoto2025adjusting}.

\subsection{Baselines and pseudo-anomaly construction}

We compare against \emph{Oracle-Selected}, \emph{Fixed-Selected}, \emph{Random-Selected}, \emph{Equal}, \emph{Pseudo-AUC-Selected}, and \emph{Bound-Selected}. \emph{Oracle-Selected} uses ground-truth development performance, while \emph{Fixed-Selected} chooses one candidate based on mean development performance across splits. \emph{Random-Selected} averages 1000 random candidate selections, and \emph{Equal} uniformly averages all candidates. \emph{Pseudo-AUC-Selected} and \emph{Bound-Selected} select candidates using pseudo-AUC and the anomaly-free bound, respectively. The proposed approach instead continuously optimizes the ensemble weights using the anomaly-free bound.
\par
For computing the bounds and pseudo AUCs, we use the pseudo-anomaly constructions introduced in \cite{wilkinghoff2026can}. \emph{Random} samples a Gaussian distribution in the embedding space, while \emph{Feature} combines individual feature dimensions from different normal embeddings. For sequence-based representations, \emph{Sequence} and \emph{Element} exchange embeddings between samples before sequence aggregation. \emph{Cross-Domain}, \emph{Cross-Class}, and \emph{Cross-Attribute} use normal embeddings from different domains, semantic classes, and attributes, respectively. The cross-* constructions require the corresponding metadata and can combine normal data across machine-type splits, whereas \emph{Random}, \emph{Feature}, \emph{Sequence}, and \emph{Element} use only normal data from the current split.

\section{Results}
\begin{table*}[t]
\centering
\vspace{-6.2pt}
\caption{
Ablation of pseudo-anomaly construction. Entries report the average
performance across DCASE~2022--2025.
}
\label{tab:pseudo_anomaly_ablation}

\setlength{\tabcolsep}{3pt}
\begin{adjustbox}{max width=0.91\textwidth}
\begin{NiceTabular}{l*{10}{c}}
\toprule

&
&
\multicolumn{3}{c}{\ac{nn}}
&
\multicolumn{3}{c}{\ac{nn} + \ac{ldn}}
&
\multicolumn{3}{c}{\ac{nn} + \ac{ldn} + VarMin}
\\

\cmidrule(lr){3-5}
\cmidrule(lr){6-8}
\cmidrule(lr){9-11}

Pseudo-Anomaly
&
requires metadata
&
P-AUC Sel.
&
Bound Sel.
&
Bound Opt.
&
P-AUC Sel.
&
Bound Sel.
&
Bound Opt.
&
P-AUC Sel.
&
Bound Sel.
&
Bound Opt.
\\

\midrule

Random & --
& 58.83  
& 58.54  
& 59.60  
& 60.61  
& \underline{63.17}  
& 63.88  
& 60.91  
& \pmb{63.81}  
& 64.34  

\\

Sequence & --
& 57.93  
& 58.15  
& 60.12  
& 60.00  
& 60.25  
& 63.80  
& 61.69  
& 61.09  
& 64.25  

\\

Feature & --
& 59.96  
& 59.17  
& \underline{61.05}  
& 61.38  
& \pmb{63.41}  
& \pmb{64.41}  
& 61.04  
& \underline{63.74}  
& \pmb{64.89}  

\\

Element & --
& 58.54  
& 58.25  
& 59.98  
& 60.40  
& 60.47  
& 63.61  
& 61.77  
& 61.02  
& 64.17  

\\

Cross-Domain & \checkmark
& 59.05  
& 59.72  
& 60.73  
& 60.53  
& 62.55  
& \underline{64.30}  
& 63.07  
& 63.40  
& \underline{64.66}  

\\

Cross-Class & \checkmark
& 60.36  
& 60.57  
& 60.78  
& 61.52  
& 61.69  
& 62.97  
& 62.69  
& 62.78  
& 63.10  

\\

Cross-Attribute & \checkmark
& \underline{60.42}  
& \pmb{61.44}  
& 60.61  
& \underline{61.90}  
& 62.42  
& 63.86  
& \underline{63.42}  
& 63.09  
& 64.48  

\\

Diverse (Global Bound) & \checkmark
& \pmb{61.39}  
& 60.82  
& \pmb{61.16}  
& \pmb{62.48}  
& 62.17  
& 63.08  
& \pmb{63.87}  
& 63.22  
& 63.56  

\\

Diverse (Mean of Bounds) & \checkmark
& --  
& 61.06  
& 60.83  
& --  
& 62.84  
& 63.40  
& --  
& 63.39  
& 63.91  

\\

Diverse (Weighted Mean of Bounds) & \checkmark
& --  
& \underline{61.18}  
& 60.79  
& --  
& 62.56  
& 64.01  
& --  
& 63.40  
& 64.62  

\\

\bottomrule
\end{NiceTabular}
\end{adjustbox}
\end{table*}

\subsection{Anomaly-free ensemble optimization}
\label{sec:ensemble_results}

\Cref{tab:ensemble_main} compares the proposed anomaly-free ensemble
optimization with several model-selection baselines. Values in brackets
denote 95\% CIs for the corresponding differences. Notably, the
\emph{Equal} ensembling baseline is strong despite simply averaging all
candidate models. Nevertheless, the proposed approach significantly
outperforms it. On average, the gains over \emph{Equal} are 1.31 percentage
points for plain \ac{nn} scores ([0.81, 1.91]), 0.83 points for
\ac{nn}+\ac{ldn} ([0.38, 1.18]), and 0.80 points for
\ac{nn}+\ac{ldn}+\ac{varmin} ([0.46, 1.16]). Since the optimization is
initialized with equal weights, these gains demonstrate that the AUC-bound objective can improve upon the equal-weight starting point.

\par
The best-performing practical selection baselines are \emph{Pseudo-AUC-Selected} and \emph{Fixed-Selected}, with one relying on metadata and the other requiring labeled anomalous samples. For \ac{nn}+\ac{ldn}, the proposed approach improves over \emph{Pseudo-AUC-Selected} by 1.93 percentage points [1.05, 2.93] and over \emph{Fixed-Selected} by 0.89 points [0.24, 1.57]. For \ac{nn}+\ac{ldn}+\ac{varmin}, the corresponding gains are 1.05 points [-0.16, 2.36] and 0.67 points [0.07, 1.26], respectively. The \ac{nn}+\ac{ldn} gains are statistically significant for both comparisons, while the \ac{nn}+\ac{ldn}+\ac{varmin} comparison with \emph{Pseudo-AUC-Selected} remains uncertain. Notably, the proposed approach outperforms \emph{Fixed-Selected} for both LDN-based settings despite its access to ground-truth anomaly labels. Thus, the proposed approach achieves strong or superior performance without the labeled anomalies or metadata required by these selection baselines.

\par
For plain \ac{nn} scoring, the proposed approach performs 0.34 percentage points below \emph{Pseudo-AUC-Selected} [-1.54, 0.82] and 0.79 points below \emph{Fixed-Selected} [-1.79, 0.42]. While the optimization does not provide a performance advantage in this setting, it still achieves 61.05\% without anomalous data or metadata. The larger gains with \ac{ldn} suggest that the score representation matters for optimizing the moment-based AUC bound, with \ac{ldn} reducing variations in absolute distance scale through the local reference density.

\subsection{Pseudo-anomaly construction}
\label{sec:pseudo_anomaly_results}

\Cref{tab:pseudo_anomaly_ablation} investigates how pseudo-anomaly construction influences anomaly-free model selection and continuous ensemble optimization. For a fixed pseudo-anomaly construction, continuous optimization generally improves over bound-based selection, most consistently for the \ac{ldn}-based scoring paradigms, where \emph{Bound Opt.} outperforms \emph{Bound Sel.} for every individual construction. It also frequently outperforms pseudo-AUC selection for the same pseudo-anomaly construction. Overall, continuous bound optimization is less sensitive to the pseudo-anomaly construction than either selection strategy. In particular, the label-free \emph{Feature} construction performs well: for \ac{nn}+\ac{ldn} and \ac{nn}+\ac{ldn}+VarMin, it achieves 64.41\% and 64.89\%, respectively, outperforming all metadata-dependent \emph{Diverse} variants. These results indicate that the proposed self-optimization approach requires neither metadata nor samples from other classes, attributes, or domains. For plain \ac{nn} scores, the gains from continuous optimization and bound aggregation are less pronounced, as also observed in \cref{sec:ensemble_results}. Nevertheless, \emph{Feature} reaches 61.05\%, close to the 61.39\% of the metadata-dependent \emph{Diverse} pseudo-AUC baseline.

\par

For the \emph{Diverse} construction, which combines multiple pseudo-anomaly sets with potentially different score scales, we also evaluate the mean and weighted mean of their individual bounds rather than a single bound over the combined pseudo-anomalies, with the weights learned using a softmax parameterization. Weighted bound aggregation benefits the \ac{ldn}-based settings, while aggregation is detrimental for plain \ac{nn} scores, where the global bound performs best. Overall, \emph{Feature} is a strong choice, achieving higher performance for the \ac{ldn}-based systems and essentially matching the best \emph{Diverse} variant for plain \ac{nn} without requiring additional metadata or cross-class, cross-attribute, or cross-domain samples.

\subsection{Effect of learnable pseudo-outlier score scaling}
\begin{table}[t]
\centering
\vspace{-6.2pt}
\caption{
Ablation of learnable pseudo-outlier score scaling. Entries report the average performance across DCASE~2022--2025.
}
\label{tab:scaling_ablation}

\setlength{\tabcolsep}{3pt}
\begin{adjustbox}{max width=\columnwidth}
\begin{NiceTabular}{l c >{\centering\arraybackslash}p{2.8cm} >{\centering\arraybackslash}p{2.8cm} >{\centering\arraybackslash}p{2.8cm}}
\toprule

Pseudo-Anomaly
&
Scaling
&
\ac{nn}
&
\ac{nn} + \ac{ldn}
&
\ac{nn} + \ac{ldn} + VarMin

\\

\midrule

Random & --
& 59.17  
& \pmb{64.11}  
& \pmb{64.46}  

\\

Random & \checkmark
& \pmb{59.60}  
& 63.88  
& 64.34  

\\

\midrule

Sequence & --
& 59.59  
& 63.71  
& 64.00  

\\

Sequence & \checkmark
& \pmb{60.12}  
& \pmb{63.80}  
& \pmb{64.25}  

\\

\midrule

Feature & --
& 60.35  
& 64.11  
& 64.56  

\\

Feature & \checkmark
& \pmb{61.05}  
& \pmb{64.41}  
& \pmb{64.89}  

\\
\midrule

Element & --
& 59.35  
& 63.01  
& 63.50  

\\

Element & \checkmark
& \pmb{59.98}  
& \pmb{63.61}  
& \pmb{64.17}  

\\
\midrule

Diverse (global) & --
& 60.60  
& 62.79  
& \pmb{63.57}  

\\

Diverse (global) & \checkmark
& \pmb{61.16}  
& \pmb{63.08}  
& 63.56  

\\
\midrule

Diverse (weighted mean) & --
& \pmb{60.93}  
& \pmb{64.33}  
& \pmb{64.74}  

\\

Diverse (weighted mean) & \checkmark
& 60.79  
& 64.01  
& 64.62  

\\

\bottomrule
\end{NiceTabular}
\end{adjustbox}
\end{table}

\label{sec:scaling_ablation}

\Cref{tab:scaling_ablation} examines the effect of learnable pseudo-outlier score scaling across pseudo-anomaly constructions and scoring paradigms. For \emph{Sequence}, \emph{Feature}, and \emph{Element}, scaling consistently improves optimization across all three scoring paradigms. For \emph{Feature}, for example, performance increases from 60.35\% to 61.05\% for \ac{nn}, from 64.11\% to 64.41\% for \ac{nn}+\ac{ldn}, and from 64.56\% to 64.89\% for \ac{nn}+\ac{ldn}+\ac{varmin}. The metadata-dependent \emph{Diverse (global)} construction shows the same trend for \ac{nn} and \ac{nn}+\ac{ldn}, while having essentially no effect for \ac{nn}+\ac{ldn}+\ac{varmin}. Thus, scaling benefits several substantially different pseudo-anomaly constructions, with the strongest gains for plain \ac{nn} scores, and is not specific to the metadata-free \emph{Feature} construction.
\par
Two cases behave differently. For \emph{Random}, scaling provides little benefit for \ac{nn} and slightly reduces performance for the \ac{ldn}-based
variants. The \emph{Random} pseudo-anomalies may have low score variance due to distance concentration in the high-dimensional embedding space, limiting the benefit of additional scaling. For the Diverse weighted-mean aggregation, scaling instead consistently reduces performance across all three scoring paradigms. A possible explanation is that scaling changes the relative magnitudes of the individual bounds and thus the resulting aggregation.

\section{Conclusion and Future Work}

We introduced anomaly-free self-optimization of \ac{asd} systems by turning
an AUC-derived bound into a directly optimizable objective. In contrast to
previous bound-based model selection, the proposed approach directly learns
continuous system parameters, with ensemble weights as a concrete example.
The learnable pseudo-outlier score scaling further adapts the surrogate pseudo-anomaly distribution to the optimization, while requiring neither anomalous
development data nor metadata-dependent pseudo-anomalies. On DCASE~2022--2025, the resulting ensembles improve over strong anomaly-free baselines and are less sensitive to pseudo-anomaly construction than conventional selection strategies. These results demonstrate that anomaly-free bounds can serve not only as model-selection criteria, but also as an optimization signal for automatically adapting \ac{asd} systems without anomalous validation data.
\par
A limitation of anomaly-free self-optimization is the unavoidable mismatch between the surrogate bound and the unknown distribution of real anomalies. Although our results show that even simple pseudo-anomaly constructions can provide an effective optimization signal, pseudo-AUC-based selection can still outperform the proposed bound in some settings. Understanding when and why direct bound optimization is less effective, and how this can be mitigated using only normal data, is an important direction for future work. Another direction is to extend the framework beyond ensemble weights to other \ac{asd}-system parameters, such as pooling and representation parameters, enabling broader anomaly-free adaptation of the \ac{asd} pipeline.

\section{Generative AI disclosure}
Generative AI tools were used for language editing and polishing of the manuscript. All scientific content, interpretations, and conclusions are the responsibility of the authors.

\clearpage
\newpage
\bibliographystyle{IEEEbib-abbrev}
\bibliography{mybib}

\end{document}